\def\year{2018}\relax
\documentclass[letterpaper]{article} 
\usepackage{aaai18}  
\usepackage{times}  
\usepackage{helvet}  
\usepackage{courier}  
\usepackage{url}  
\usepackage{graphicx}  
\usepackage{color}
\usepackage{amsmath}
\usepackage{amsfonts}
\usepackage{amssymb}

\usepackage[symbol]{footmisc}

\usepackage{algorithm}
\usepackage{algorithmic}

\newcommand{\ep}{\mathbb{E}}
\newcommand{\bb}[1]{\mathbb{#1}}

\newcommand{\tabincell}[2]{\begin{tabular}{@{}#1@{}}#2\end{tabular}}

\begin{document}
\title{Understanding Human Behaviors in Crowds by \\ Imitating the Decision-Making Process}
\author{Haosheng Zou,  Hang Su,  Shihong Song,  Jun Zhu\thanks{Corresponding author. The work was supported by the National NSF of China Projects (Nos. 61571261, 61620106010, 61621136008).}\\
Dept. of Comp. Sci. \& Tech., State Key Lab of Intell. Tech. \& Sys., TNList Lab, CBICR Center\\
Tsinghua University, Beijing, China\\
 \texttt{\{zouhs16@mails, suhangss@mail, songsh15@mails, dcszj@mail\}.tsinghua.edu.cn}
}
\maketitle

\begin{abstract}
Crowd behavior understanding is crucial yet challenging across a wide range of applications, since crowd behavior is inherently determined by a sequential decision-making process based on various factors, such as the pedestrians' own destinations, interaction with nearby pedestrians and anticipation of upcoming events. In this paper, we propose a novel framework of Social-Aware Generative Adversarial Imitation Learning (SA-GAIL) to mimic the underlying decision-making process of pedestrians in crowds. Specifically, we infer the latent factors of human decision-making process in an unsupervised manner by extending the Generative Adversarial Imitation Learning framework to anticipate future paths of pedestrians. Different factors of human decision making are disentangled with mutual information maximization, with the process modeled by collision avoidance regularization and Social-Aware LSTMs. Experimental results demonstrate the potential of our framework in disentangling the latent decision-making factors of pedestrians and stronger abilities in predicting future trajectories.
\end{abstract}

\section{Introduction}

\begin{figure}[!thb]
\centering
\includegraphics[width=0.9\linewidth]{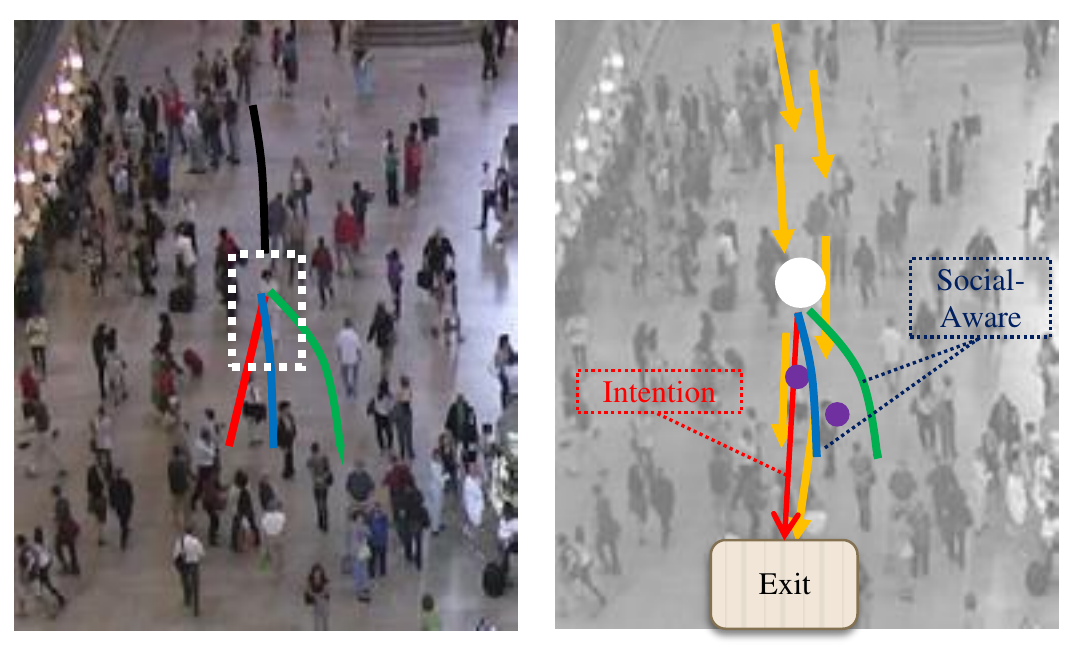}
\caption{
Left: Illustration of an imitating decision-making process. The dashed white rectangle indicates the pedestrian of interest. The black segment is her observed trajectory. Controlled by a latent code, the process generates different future trajectories in red, green and blue.
Right: Data of sufficient pedestrians reveal the local approximate motion trend (yellow arrows) towards the exit in the bottom left corner. We infer this exit to be the intention of this pedestrian (denoted white disc). However, to accommodate for nearby pedestrians (purple discs), other assignments of the latent code generate more social-aware trajectories.
}
\label{fig:intro}
\end{figure}
With the increase of population and diversity of human activities, accurate analysis of human behaviors in crowds becomes an essential requirement across a wide range of domains, e.g., visual surveillance \cite{Scene_Shao_2014,Deeply_Shao_2015}, robot navigation in crowds \cite{robicquet2016learning}, etc. Recently, there has been a growing interest in developing computational methodologies for modeling crowd behaviors, including visual representations~\cite{zhou2011random,luber2010people,Crowd_Zhou_2012,Tracking_Rodriguez_2009}, detection of abnormal behaviors~\cite{mehran2009abnormal} and modeling interaction among pedestrians~\cite{andriluka2012human}. Despite these strong needs, the complexities of crowded scenes such as frequent occlusions among pedestrians and large variations of human behaviors across space-time severely impede the implementation of conventional algorithms. A lot of efforts have been made to address these issues by leveraging the recent advances in deep learning~\cite{su2016crowd,su2017forecast,alahi2016social,yi2016pedestrian,Lee2017DESIRE}. However, these methods only learn the general historical motion statistics of a scene, without deeper investigation into humans in crowds.

Without more thorough modeling, learning crowd behaviors from videos remains difficult partly because the individual motion is affected by many factors that are not directly captured by videos.
Human behaviors in crowds are inherently determined by a sequential decision-making process according to the persons of interest, environments and social rules~\cite{ali2008taming,Xie2017Learning}. In general, pedestrians first decide their destinations, pick paths by considering the neighboring crowd state, and constantly make minor modifications to follow the majority or avoid collision (see Fig. \ref{fig:intro} (Right) for example). This cognitively-based method was popularly used in crowd simulation previously~\cite{yu2007decision,pellegrini2009you}. Nevertheless, only a few attempts have been made on vision-based analysis, which resort to simplified rules, energy functions~\cite{pellegrini2009you,Xie2017Learning} and grid-based planning \cite{ziebart2009planning-based}, with large amounts of hand-crafted design and specification of dynamics and features.

Admittedly, inferring the aforementioned decision-making process of pedestrians is challenging since it is complex and determined by various factors, including personal intentions, neighboring pedestrians, and physical obstacles~\cite{pellegrini2009you}. The interaction with neighboring pedestrians has been modeled by social pooling layers~\cite{alahi2016social,Lee2017DESIRE} and coherently regularized RNNs~\cite{Cherent_Zhou_2012,su2016crowd}. However, the pedestrian's own intention, though plays a key role as the internal driving force of human behaviors~\cite{ziebart2009planning-based,intent_Monfort_2015,Xie2017Learning}, remains to be better modeled. 

In addition to the quantities of the decision-making factors, the way that humans consider these factors before making decisions is also not well exploited. Humans do not simply act \emph{responsively} to them. They also \emph{anticipate} future states and induce the current action backwards. One typical example is when two people travel in orthogonal directions, they constantly locate each other to forecast whether they will collide.
This innate ability of humans to simulate what is about to happen before taking actions is missing in almost all recent works. 

\subsection{Our proposal}

Different from previous algorithms based on the statistics of visual appearance, we propose a novel Social-Aware Generative Adversarial Imitation Learning (SA-GAIL) framework to understand crowd behaviors by inferring the decision-making process of pedestrians in crowds. Specifically, we infer the humans' decision-making process by anticipating their future paths as illustrated in Fig. \ref{fig:intro}, which requires a deeper understanding of the behaviors in determining not only what the activity is but also how the activity should be unfolded. To this end, we imitate human behaviors from the observed trajectories by extending the recent framework of Generative Adversarial Imitation Learning (GAIL)~\cite{ho2016generative}, which is capable of training generic neural network policies to produce expert-like movement patterns from limited demonstrations.

Compared with existing supervised learning methods~\cite{su2016crowd,su2017forecast,alahi2016social,yi2016pedestrian,Lee2017DESIRE}, Imitation Learning (IL), especially GAIL, fits in more naturally for the human decision-making process~\cite{li2017inferring} since it enjoys both a sound theoretical analysis and efficient training.
In particular, the numerous human trajectories in the training set are regarded as expert demonstrations, which reflect the human decision-making process in a crowded scenario. Following GAIL, we devise a generator (policy) based on the recurrent encoder-decoder framework which tries to generate behaviors matching the expert demonstrations, and a discriminator which tries to distinguish the generated trajectories from expert demonstrations. The generator and discriminator are jointly optimized as playing an adversarial minimax game.

To properly model the pedestrians' latent intentions, 
it is natural to first disentangle the latent intention from other social factors. To address this, we propose to learn semantically meaningful latent codes that reflect different factors of the decision-making process. Specifically, the objective function of GAIL~\cite{ho2016generative} is further augmented with a mutual information term between the latent codes and the generated paths \cite{chen2016infogan,li2017inferring}. This modification is based on the observation that pedestrians with the same entrance and destination may still behave differently, because they face different neighborhood conditions. The latent codes thus provide an elegant way to represent such diversity, with one setting expectedly corresponding to the policy to walk straight to the destination, and others corresponding to the influence of neighbors, as illustrated by the outputs in Fig. \ref{fig:intro} (Left). 
It's noteworthy that the disentangled latent representations are learned in an \emph{unsupervised} manner without any human labeling of the different factors. This is different from the traditional physics-inspired Social Force model~\cite{helbing2000simulating,mehran2009abnormal}, where different forces driving pedestrians moving are defined in a hand-crafted fashion.

We further imitate the human decision-making process by introducing human's ability of world simulation into SA-GAIL. Here, we mainly consider the instinct of collision avoidance. Inspired by the ORCA framework \cite{van2011reciprocal} in crowd simulation, we propose an original communication-simulation mechanism across pedestrians. Each pedestrian propagates his/her desired candidate action to nearby pedestrians, and only executes it when no collision is detected. 
The pedestrians will make an efficient move to avoid pending collisions in advance with minimal effort. In this paper, we assume that pedestrians choose the paths with a minimal amount of movement and turning effort upon interacting with each other, which behaves like people observing neighbors and acting accordingly.

Extensive experiments demonstrate that SA-GAIL can not only understand the present crowd behaviors but also predict their future paths after penetrating into the underlying decision-making process. To summarize, our main contributions are:
\begin{enumerate}
\item
We present a first attempt to apply a novel data-driven Imitation Learning framework to model the sequential decision-making process of pedestrian behaviors.
With the imitation process, we reason about and model crowd motion from a perspective more similar to human decision-making.
\item The different factors of pedestrian decision-making are disentangled in an unsupervised manner with mutual information maximization. We specifically infer the human intentions apart from human-human interactions, endowing the policy with certain interpretability.
\item We bring the idea of world simulation into behavior understanding by introducing a collision avoidance regularization which does not jeopardize neural network training. Our model behaves more human-like and avoids predicting intuitively unreasonable paths.
\end{enumerate}

\begin{figure*}[!thb]
\centering
\includegraphics[width=0.95\linewidth]{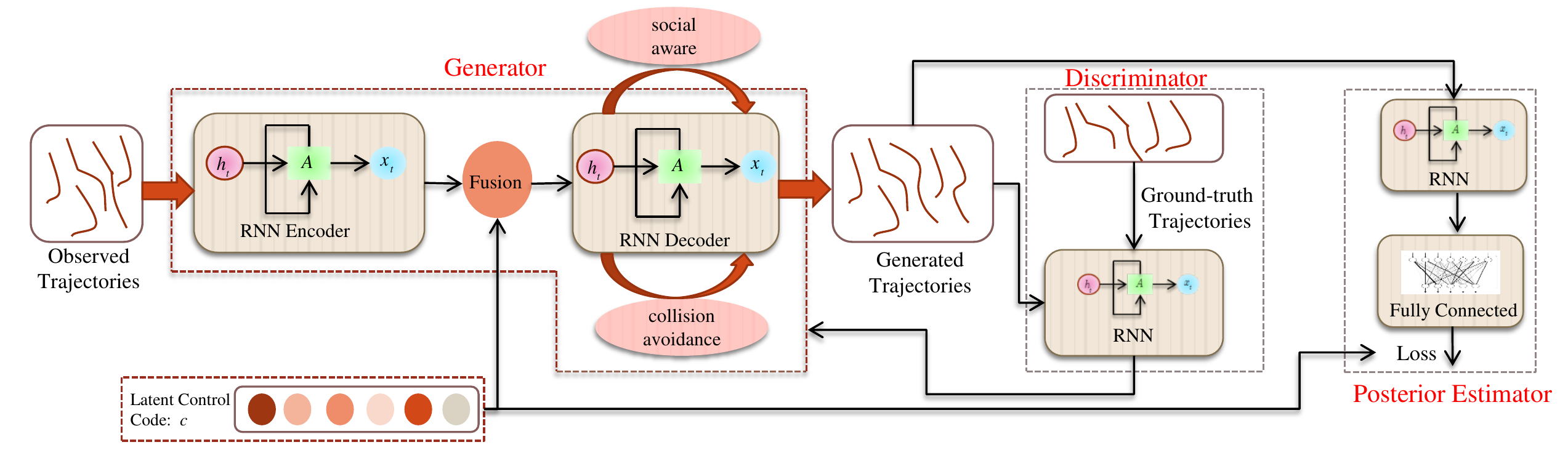}
\caption{Overview of our proposed SA-GAIL. Observed trajectories are fed into the policy to generate future paths. The policy/generator adopts an encoder-decoder architecture where latent codes could be fused with the decoder, augmented with a collision avoidance regularization and social awareness. The generated paths, concatenated with the observed ones, are then fed into the discriminator to guide the generator training and into the posterior to recover the latent code.}
\label{fig:method}
\end{figure*}

\section{Methodology}

To model the decision-making process of pedestrian behaviors, we propose SA-GAIL for crowd behavior understanding. In this section, we first formulate the problem and overview the general framework of SA-GAIL; then, we describe each component of the architectures;  finally we introduce the optimization techniques for trajectory prediction and intention inference, respectively.

\subsection{Problem Formulation}
\label{sec:formulation}

We implement behavior understanding via motion prediction, i.e., given an observed trajectory of length $T_1$, forecast the future trajectory of length $T_2$. The trajectories are already obtained from tracking in the format of sequences of coordinates, with the observed trajectory of pedestrian $i$ being $X_i = \{x_1, x_2, ..., x_{T_1}\}$ and his/her ground-truth future trajectory being $Y_i = \{y_{T_1 + 1}, y_{T_1 + 2}, ..., y_{T_1 + T_2}\}$. Each $x_i$ and $y_i$ is a two-dimensional coordinate in the video image. Instead of treating trajectory prediction as sequence-to-sequence mapping \cite{su2016crowd,su2017forecast,alahi2016social,Lee2017DESIRE}, we tackle it from a decision-making perspective. Borrowing the notion of states and actions in MDP (Markov Decision Process), we have a policy generating actions $\hat{Y}_i$ conditioned on previous states, where states correspond to previous coordinates of the trajectory and actions correspond to the pedestrian's next position. In addition to generating one trajectory per person as existing works do \cite{su2016crowd,su2017forecast,alahi2016social,yi2016pedestrian}, we also generate multiple trajectories to visualize intention inference.

\subsection{System Overview}

As shown in Figure \ref{fig:method}, SA-GAIL has three constituent networks, namely the policy/generator $G$, the discriminator $D$ and the variational posterior estimator $Q$. The three networks are jointly optimized with Imitation Learning (IL) to deconstruct the underlying decision-making process.

Specifically, the policy $G$ takes as input the observed trajectories and generates future trajectories, which we parameterize as a Recurrent Neural Network (RNN). It generates actions in an auto-regressive manner:
\footnotesize
$$\hat{y}_t = G(x_1, ..., x_{T_1}, \hat{y}_{T_1 + 1}, ..., \hat{y}_{t - 1}), \forall T_1 < t \leq T_1 + T_2.$$
\normalsize
\noindent When trained to perform intention inference, we also inject a latent code $c$ to the policy to control its output:
\footnotesize
$$\hat{y}_t = G(c; x_1, ..., x_{T_1}, \hat{y}_{T_1 + 1}, ..., \hat{y}_{t - 1}), \forall T_1 < t \leq T_1 + T_2.$$
\normalsize
We design three Social-Aware components into $G$:
\begin{itemize}
\item Intention inference with latent code
\item Collision avoidance regularization with simulation
\item Social-Aware LSTMs for human-human interaction
\end{itemize}
\noindent The obtained full trajectories from $G$, merging the observed and generated parts, are then fed to 1) the discriminator $D$ to output a score and 2) the posterior $Q$ to try to recover the latent code injected to the generator before policy execution.

The generator and discriminator are jointly optimized with GAIL \cite{ho2016generative} in the form of an adversarial minimax game as GAN \cite{goodfellow2014generative}:
\footnotesize
$$\min_G \max_D \ep \big[\log D([X_i, Y_i]) \big] + \ep \big[\log (1 - D([X_i, \hat{Y}_i])) \big].$$
\normalsize

\noindent The posterior estimator is further optimized together with the generator to maximize the mutual information \cite{chen2016infogan} between the generated paths and the injected code:
\footnotesize
\begin{equation*}
I([X_i, \hat{Y}_i]; c) = H(c) - H(c | [X_i, \hat{Y}_i]), 
\end{equation*}
\normalsize
where $H(\cdot)$ is entropy. The policy will then be controlled by the latent code to generate interpretable trajectories to some extent as performing intention inference.

\subsection{Generator/Policy Design}

The generator, acting as the policy for the decision-making process, should intuitively grasp a fair understanding of the previous states to decide its next action. 
However, the sequential nature of pedestrian trajectories poses practical challenges for conventional feed-forward neural policies \cite{ho2016generative,li2017inferring} to reason about the status of each individual as well as the dynamic interaction between them.

Towards this end, we propose the generator $G$ with an architecture inspired by the encoder-decoder model~\cite{cho2014learning,Sequence_Sutskever_2014} to account for the sequential nature, as depicted by the Generator part in Figure~\ref{fig:method}. To capture the individual status from the observed trajectory, we input the observed part to the encoder RNN one coordinate per timestep, which progresses as
\footnotesize
$$h_t = RNN_{enc}(x_t, h_{t - 1}), \forall 1 \leq t \leq T_1.$$
\normalsize
\noindent The last hidden state, $h_{T_1}$, is treated as the fixed-length descriptor of the observed part of the trajectory to initialize the decoder RNN. Useful information about the pedestrian status such as his/her own navigation style and desired destination is extracted in this fixed-length vector.

The decoder RNN then takes $h_{T_1}$ as its initial hidden state and $x_{T_1}$ as its first input. By recursively inputting its last generated action to itself, the decoder RNN evolves as
\footnotesize
\begin{eqnarray*}
\hat{y}_{T_1 + 1}, h_{T_1 + 1} & = & RNN_{dec}(x_{T_1}, h_{T_1}) \\
\hat{y}_t, h_t & = & RNN_{dec}(\hat{y}_{t - 1}, h_{t - 1}),\\
&& \forall T_1 + 1 < t \leq T_1 + T_2. 
\end{eqnarray*}
\normalsize
We then build the three Social-Aware components into this encoder-decoder architecture. We keep the encoder RNN simple to capture only the individual status, and augment the decoder RNN with the Social-Aware extensions.

\subsubsection{Intention Inference:}
\label{sec:intent}
In this work, for simplicity we consider the two most critical factors underlying the pedestrian decision-making process, namely the latent intention of the pedestrians themselves and the interaction with nearby pedestrians. 
As a result of the two factors, in a sparse environment with few people ahead people usually walk straight to their intented destinations, and in crowded scenarios where there are many people around travelling in different directions, people have to avoid collision, respect others' personal space and thus take detours deviating from the ideal straight path.
Therefore, two pedestrians sharing similar observed trajectories may still continue to navigate differently due to the different crowd states they face. The assumption that people sharing resembling observed paths would continue behaving similarly, upon which the recent deep learning methods are based \cite{su2016crowd,su2017forecast,alahi2016social,yi2016pedestrian,Lee2017DESIRE}, is thus rather debatable, and a fairer assumption would be that they share similar latent intentions instead of similar future behavior.

\begin{figure}
\centering
\includegraphics[width=1\columnwidth]{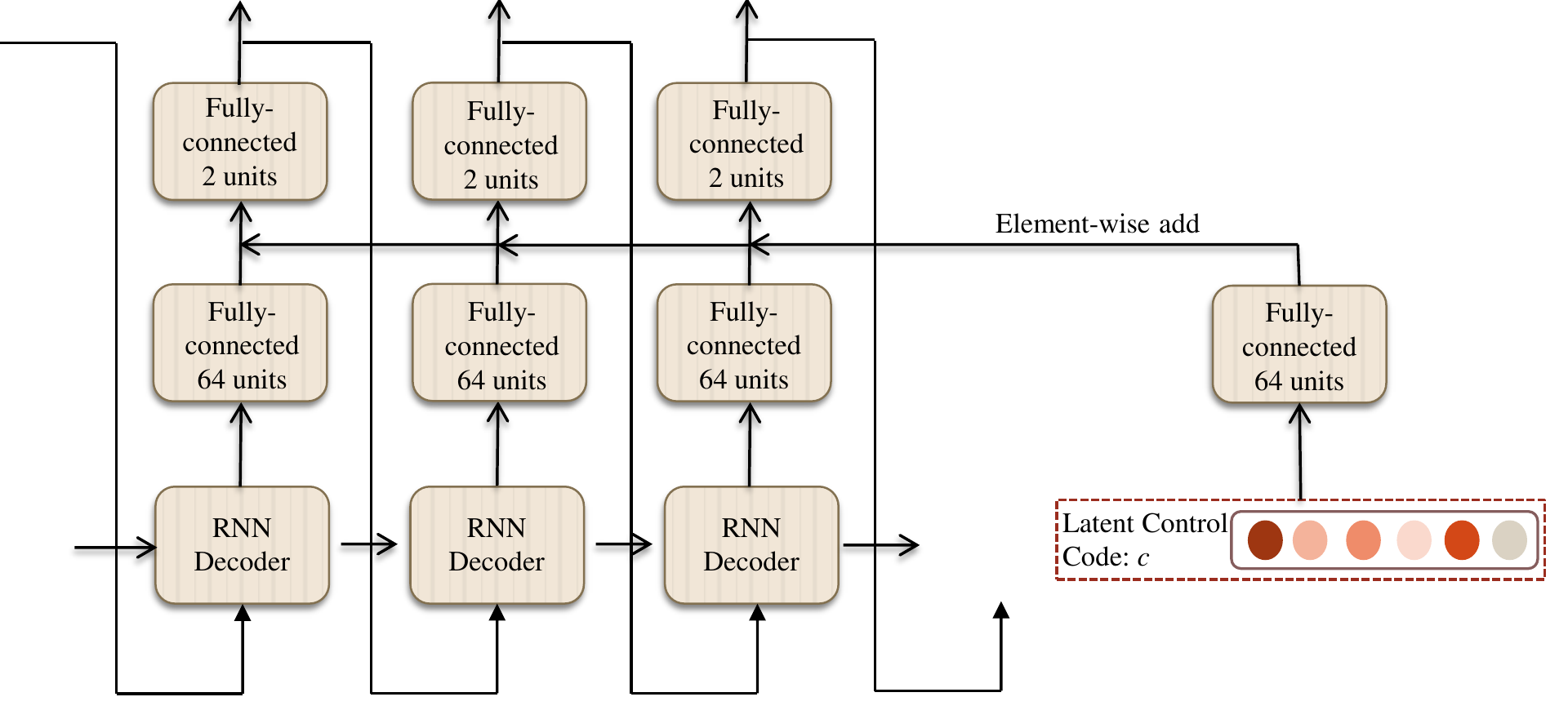}
\caption{An example of the decoder RNN with injected code. The code is first embedded to a higher dimension equal to one of the hidden layers, and is then element-wise added to the hidden layer activation.}
\label{fig:decoder}
\end{figure}

With sufficient data of pedestrian behavior, we assume that both types of future behavior (ideal straight and deviating detours) are of significant presence. Furthermore, even though deviating detours may exhibit wide diversity, the ideal straight trajectory is generally stable and direct to destination, forming a generalized ``mode'' conditioned on observed trajectories. This provides the 
possibility of learning the mode in a data-driven fashion to gain a deeper understanding of the decision-making process. To model and disentangle the latent intention apart from the diversity caused by the interaction factor, we propose to inject an additional code $c$ into our policy so that the policy could act differently controlled by the code. More specifically, the code is injected after the encoder RNN based on the intuition that similar observed trajectories should result in similar descriptive vectors $h_{T_1}$ from the encoder RNN. We fuse $c$ with $h_{T_1}$ in a certain way so that different $c$ causes the decoder RNN to generate different future paths, of which an example architecture is shown in Fig. \ref{fig:decoder}. The relationship between $c$ and genenrated $\hat{Y}_i$ is discovered in an \emph{unsupervised} manner without any further labeling, as will be detailed in Sec. \ref{sec:opt}.

\subsubsection{Collision Avoidance:}

A policy imitating ordinary pedestrians' decision-making should naturally prevent itself from generating two crashing trajectories, as ordinary people simulate, foresee collision and then act accordingly. Recent deep learning methods achieve this only \emph{implicitly} through mean squared error of each trajectory w.r.t. its ground-truth \cite{su2016crowd,su2017forecast,alahi2016social,yi2016pedestrian,Lee2017DESIRE}.
To explicitly incorporate collision avoidance into our generator, we draw inspiration from ORCA (Optimal Reciprocal Collision Avoidance) \cite{van2011reciprocal} and propose a novel regularization mechanism of instant simulation into our policy.
In ORCA, each agent senses its nearby agents' current velocities and maps its own to a velocity that's guaranteed to be safe for a time window, indirectly simulating for that time. The mapping operation is implemented as an inner optimization loop of linear programming, thus not directly applicable with neural networks' backpropagation. 
We accordingly propose a simplified differentiable operation in this regard. Each agent still senses all the other agents within the same mini-batch, but instead of performing velocity mapping, it will directly stop moving if its generated next action collides with one or more of the others. Therefore, the decoding RNN now performs the following recurrence:
\footnotesize
\begin{eqnarray*}
\hat{y}_t, h_t & = & RNN_{dec}(\hat{y}_{t - 1}, h_{t - 1})\\
\hat{y}_t & = & \left\{
\begin{array}{ll}
\hat{y}_{t - 1}, & \mathrm{if\ colliding} \\
\hat{y}_t, & \mathrm{if\ safe}
\end{array}
\right.
\end{eqnarray*}
\normalsize
The conditional selection of next actions is implemented in TensorFlow enabling backpropagation.
Collision is determined if two actions lie closer than several pixels to each other. The decoder RNN will continue output the same action until its generated action becomes safe. This directly causes the generated trajectories to be much shorter than and deviating from the ground-truth, thus heavily penalizing collision during training.
Compared to ORCA, our mechanism equivalently simulates for one timestep into the future (0.5 seconds at 2 fps), treats generated actions as simulated candidates and maps them directly to zero velocity if colliding.

\subsubsection{Social-Aware LSTM:}

In addition to coupling the LSTMs of each individual with collision avoidance, we further model the interaction between pedestrians inspired by \cite{alahi2016social,su2016crowd}. 
We employ an operation across RNNs similar to Social Pooling \cite{alahi2016social}, and construct a vicinity tensor $V_t^i$ of size $N \times N \times H$ in each RNN's neighborhood of pixel range $N \times N$. $H$ is the dimension of the RNN hidden state. The tensor $V_t^i$ for trajectory $i$ at timestep $t$ is computed as follows:
\footnotesize
$$V_t^i = \sum_{j \in \mathcal{N}_i} \mathbf{I}[x_t^j - x_t^i] h_{t-1}^j,$$
\normalsize
where $\mathcal{N}_i$ is the neighboring pedestrians of person $i$, $\mathbf{I}[x_t^j - x_t^i]$ is an indicator function to locate the neighbor, and $h_{t-1}^j$ is the hidden state of the RNN of person $j$ at timestep $t-1$.
This vicinity tensor is further embedded to a compact representation with the same dimension as the RNN hidden state, and added to the hidden state as \cite{su2016crowd} does.

\subsection{Discriminator and Posterior}

The distinction of our framework w.r.t. recent works \cite{su2016crowd,su2017forecast,alahi2016social,yi2016pedestrian,Lee2017DESIRE} lies in that with GAIL the supervision to the policy is provided by the discriminator. To feed the discriminator with necessary information, 
we devise the discriminator $D$ to take as input the full generated trajectory, combining the observed and generated parts. $D$ is also of a recurrent structure with RNNs first processing the whole trajectory, and the last hidden state is treated as trajectory descriptor to be passed to fully-connected layers. The fully-connected layers are followed by LeakyReLU nonlinearity as suggested by \cite{radford2015unsupervised}.
The final fully-connected layer outputs the probability of the trajectory being real as in standard GAN. 

To establish the relationship between the code $c$ and the policy $G$, a variational posterior estimator is further needed to lowerbound the mutual information \cite{chen2016infogan}. 
Similarly as the discriminator, we build the posterior taking the combined trajectories as input with output corresponding to the prior code distribution.

\subsection{Optimization Algorithms}
\label{sec:opt}

In light of the sequential decision-making process of pedestrian behavior, we propose to apply Imitation Learning (IL) and corresponding optimization algorithms to let a policy imitate crowd motion from data. The data, tracked tracklets, serve as expert demonstrations in terms of IL.

\begin{algorithm}[htp!]
   \caption{SA-GAIL}
   \label{alg:sagail}
\begin{algorithmic}
   \STATE {\bfseries Input:} Tracked trajectories $[X_i, Y_i]$; initial policy $G_0$, discriminator $D_0$ and variational posterior estimator $Q_0$
   \STATE \textbf{Output: } Learned policy $G$
   \FOR{$i=0, 1, 2, ... $}
       \STATE $\dag$ Sample a batch of latent codes from pre-specified distribution: $c_i \sim P(c)$
       \STATE Generate a batch of trajectories $\hat{Y}_i$ conditioned on $X_i$ 
       
       $\dag$ (and $c_i$ for intention inference, fixed in each rollout)
       \STATE Sample ground-truth trajectories $[X_i, Y_i]$
       \STATE Gradient descent on $D$ to minimize
       $$ \hat{\mathbb{E}}_{[X_i, \hat{Y}_i]}[\log D([X_i, \hat{Y}_i])] + \hat{\mathbb{E}}_{[X_i, Y_i]}[\log(1 - D([X_i, Y_i]))]$$
       \STATE $\dag$ Gradient descent on $Q$ to minimize
       $$ - \lambda \hat{\mathbb{E}}_{[X_i, \hat{Y}_i]} [\log Q(c|[X_i, \hat{Y}_i])]$$
       \STATE Update policy $G$ with TRPO to maximize the following reward (without intention inference):
       $$
       \hat{\bb{E}}_{[X_i, \hat{Y}_i]} [D([X_i, \hat{Y}_i])]
       $$
       \STATE $\dag$ or (for intention inference):\footnotemark[1]
       $$
       \hat{\bb{E}}_{[X_i, \hat{Y}_i]} [D([X_i, \hat{Y}_i])] + \lambda \hat{\mathbb{E}}_{[X_i, \hat{Y}_i]} [\log Q(c|[X_i, \hat{Y}_i])]
       $$
   \ENDFOR
\end{algorithmic}
\end{algorithm}

We advocate to apply GAIL \cite{ho2016generative}, sustaining the efficiency of gradient-based learning while still formulating the problem as occupancy matching as Inverse RL does. GAIL introduces a discriminator to distinguish the generated state-action pairs from the expert-demonstration ones, and this discriminator guides the learning of the policy model, drawing inspirations from GANs \cite{goodfellow2014generative}. The gradient is not directly backpropagated from $D$ to $G$, but through policy gradient algorithms such as TRPO \cite{schulman2015trust}.

\subsubsection{Motion Prediction}

For pure motion prediction, we take away the intention inference part of our framework, leaving only $G$ and $D$. The whole system, with Social-Aware LSTMs and collision avoidance, is trained with the GAIL algorithm \cite{ho2016generative}, as outlined in Algorithm \ref{alg:sagail} without intention inference (the steps beginning with $\dag$). The algorithm optimizes
\footnotesize
\begin{eqnarray*}
\min_D \max_G & \hat{\mathbb{E}}_{[X_i, \hat{Y}_i]}[\log D([X_i, G(X_i))])] \\
+ & \hat{\mathbb{E}}_{[X_i, Y_i]}[\log(1 - D([X_i, Y_i]))]
\end{eqnarray*}
\normalsize
\subsubsection{Intention Inference}

We employ the whole framework of SA-GAIL together with intention inference. The three networks, $G$, $D$ and $Q$, given the sampled latent code $c$, are jointly optimized with the SA-GAIL algorithm as fully outlined in Algorithm \ref{alg:sagail}, similar to \cite{li2017inferring,Hausman2017MultiModalIL}. The algorithm optimizes\footnote[1]{We regret that these formulas have typos in the AAAI version.}

\footnotesize
\begin{eqnarray*}
\min_{D} \max_{G, Q} & \hat{\mathbb{E}}_{[X_i, \hat{Y}_i]}[\log D([X_i, G(X_i))])] \\
+ & \hat{\mathbb{E}}_{[X_i, Y_i]}[\log(1 - D([X_i, Y_i]))] \\
+ \lambda & \hat{\mathbb{E}}_{[X_i, G(X_i)]} [\log Q(c|[X_i, G(X_i)])]
\end{eqnarray*}

\normalsize

\subsection{Characteristics of SA-GAIL}

\begin{itemize}
\item We propose the collision avoidance regularization with the idea of simulating the future. Instead of naively regularizing the objective function, it incorporates an intuitive rule into the network architecture (thus a layer) and fits in gradient-based learning (still differentiable).

\item Our generator design is novel with non-trivial analysis. We found that a sequence generation model as \cite{alahi2016social} didn't perform well on our dataset. We deduce that it's mainly because the same set of weights process both observed and generated paths, so we separate the roles of understanding pedestrian history and predicting future. With this principle, we add the three social-aware components all into the decoder, distinct from \cite{alahi2016social,su2016crowd,su2017forecast} where social-awareness is distributed all across the network.

\item Our SA-LSTM with an additive integration of vicinity and RNN states models more types of crowd motion than coherence \cite{su2016crowd,su2017forecast} and is more stable under empty vicinity than \cite{alahi2016social}. We also adopt the standard GAN discriminator because the recent WGAN discriminator exhibits more instability due to clipping on LSTM discriminator weights and unbounded output.
\end{itemize}

\section{Experiments}

We demonstrate the effectiveness of SA-GAIL on two tasks of pedestrian behavior understanding: behavior prediction and intention inference.
Behavior prediction is the task of interest in all recent works \cite{su2016crowd,su2017forecast,alahi2016social,yi2016pedestrian,Lee2017DESIRE}, which we implement as a preliminary verification of SA-GAIL. We then conduct intention inference, delving deeper into the decision-making process and seeking generalized modes across individuals for disentanglement  and interpretability.

\subsection{Dataset and Experimental Settings}

\begin{figure}[!thb]
\centering
\includegraphics[width=0.65\columnwidth]{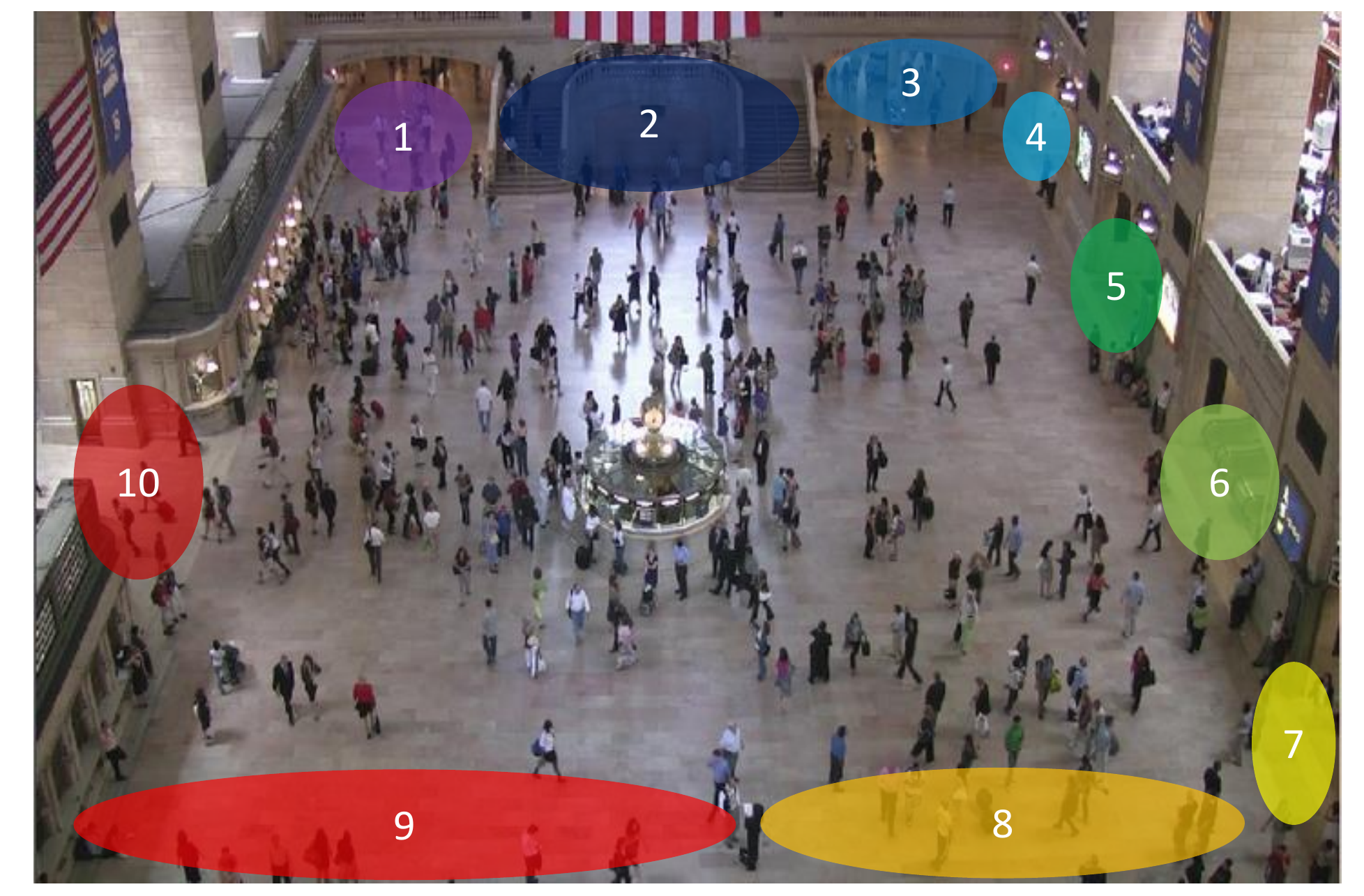}
\caption{Ten entrance/exits regions labeled in the scene.}
\label{fig:scene}
\end{figure}

We conducted all experiments on the publicly available Central Station dataset \cite{zhou2011random}, which is a surveillance video of 33 minutes long with more than 40,000 keypoint tracklets. The scene is shown in Fig. \ref{fig:scene}, with ten entrance/exit regions manually labeled.
This dataset is a highly challenging one in pedestrian motion analysis with usually over 100 people simultaneously in the unstructured scene.

As per Sec. \ref{sec:formulation}, we fix $T_1 = 9$ and $T_2 = 8$ in all our experiments. In other words, the first 9 coordinates are treated as observed part of trajectories, and our system predicts the last 8 coordinates. Shorter tracklets are ignored. We fix the lengths $T_1$, $T_2$ for simplicity, but our system is easily extended to variable-length sequences since the neural networks taking the trajectories as input are all of recurrent nature.
We sample all trajectories at a frame rate of 2 fps.
The video is 720 pixels in width and 480 pixels in height. We normalize the two dimensions of coordinates respectively w.r.t. the size so that all coordinates lie within $[0, 1]$.

We specify the basic network design as follows: we use an LSTM with 128 units for the encoder of the policy, and an LSTM with 128 units followed at each timestep by one fully-connected layer with 64 units and a final output fully-connected layer with 2 units. The hidden fully-connected layer employs ReLU nonlinearity as suggested by \cite{radford2015unsupervised} for generator. The 2-dimensional output is treated as Gaussian mean with pre-specified logstd to parameterize a stochastic policy for TRPO.
We adopt a similar architecture for the discriminator and posterior, where we use an LSTM with 128 units to process the whole sequence and add a fully-connected output layer to the last output of the LSTM. For the discriminator the output layer has only one sigmoid unit for the probability of the trajectory being real, and for the posterior a softmax distribution. We train SA-GAIL following the training procedure in \cite{ho2016generative,li2017inferring}.

\subsection{Behavior Prediction}

\begin{table}[!t]
\centering
\caption{Error of generated trajectories. The ones marked with $\dag$ are taken from \cite{yi2016pedestrian}.}
\label{tab:err}
\begin{tabular}{llll}
                                      & normADE & ADE     & FDE     \\
\hline
Constant velocity\dag                     & 5.86\%  & -       & -       \\
SFM\dag                                   & 4.45\%  & -       & -       \\
LTA\dag                                   & 4.35\%  & -       & -       \\
Behavior CNN\dag                          & 2.52\%  & -       & -       \\
\hline
Vanilla LSTM                          & 2.39\%  & 14.57 & 27.78 \\
SA-LSTM                               & 2.14\%  & 12.82 & 25.51 \\
\tabincell{r}{Vanilla LSTM with\\collision avoidance} & 2.20\%  & 13.11 & 26.19 \\
\hline
\tabincell{r}{\textbf{full SA-GAIL} (no code)}       & \textbf{1.98\%}  & \textbf{11.98} & \textbf{23.05} \\
\tabincell{r}{SA-GAIL without\\SA-LSTM}               & 2.21\%  & 13.17 & 26.55 \\
\tabincell{l}{SA-GAIL without\\collision avoidance}   & 2.09\%  & 12.61 & 25.08 \\
\tabincell{r}{SA-GAIL without GAIL}                  & 2.17\%  & 12.95 & 25.95 \\
\hline
\end{tabular}
\end{table}

For this task, we take away the posterior and code input of the networks so our system directly imitates the decision-making process and emphasizes individual prediction.

\begin{figure*}[!thb]
\centering
\includegraphics[width=1\linewidth]{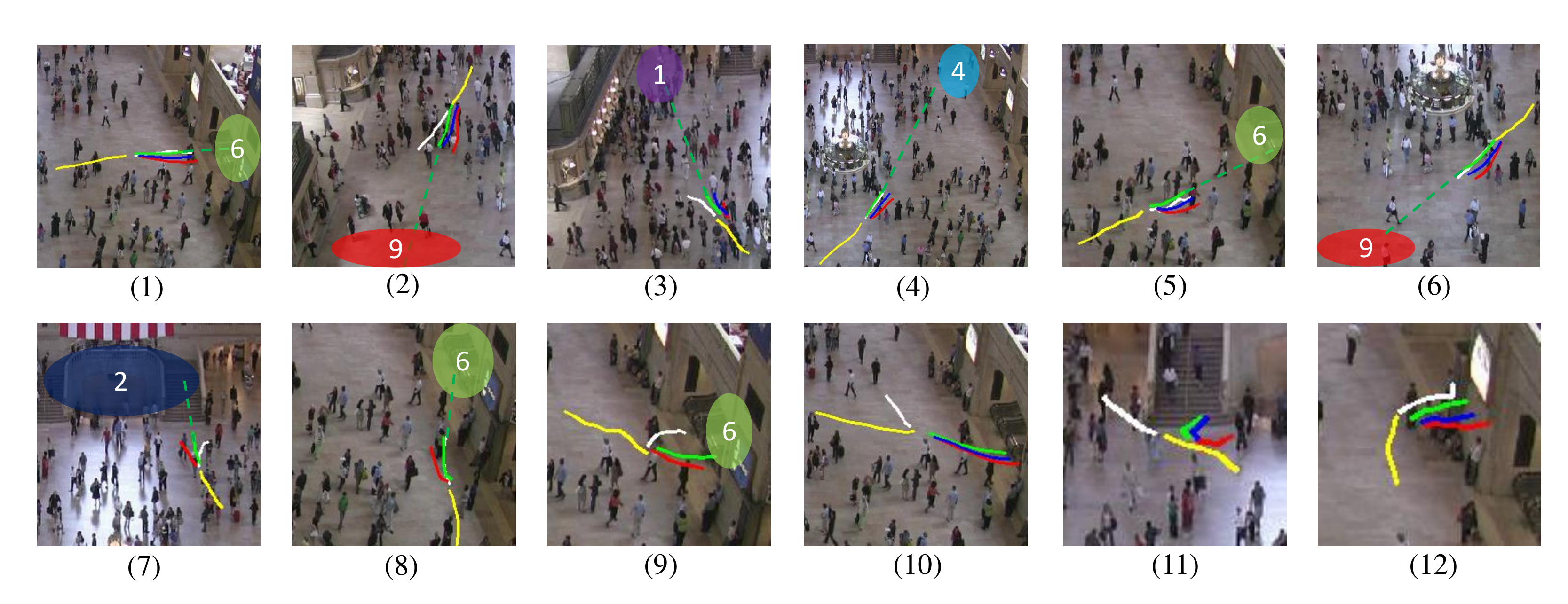}
\caption{Given observed (yellow) trajectories, our model generates different future trajectories (red, green, blue) controlled by the code, with intention orientations illustrated in dashed green. Ground-truth future paths are shown in white. (1-6): results with 3-dimensional code. (7-9): results with 2-dimensional code. (10-12): some failure cases.}
\vspace{-0.5em}
\label{fig:intent}
\end{figure*}

\textbf{Comparison methods:}
To the best of our knowledge, only  
\cite{yi2016pedestrian} has done extensive experiments on the same crowd scene, though with different settings of $T_1$, $T_2$, sampling frame rate and trajectory labeling. We refer to some of their results as approximate baselines (entries with $\dag$ in Table \ref{tab:err}). We establish three baselines of our own, namely the vanilla LSTM without any further modification, the LSTM with only vicinity tensors (SA-LSTM) and the LSTM with only collision avoidance. These methods are all trained with supervised learning and are compared to our full system (excluding the intention inference part). The contribution of each part of our system, the SA-LSTM, the collision avoidance layer and GAIL training, is demonstrated with ablation tests by removing each module.

\textbf{Evaluation metrics:}
We evalutate prediction performance with three metrics, the normalized Avarage Displacement Error (normADE), the Avarage Displacement Error (ADE) in terms of pixels and the Final Displacement Error (FDE) also in terms of pixels. The normADE is the error between generated and ground-truth coordinates normalized with respect to the image width and height, as used in \cite{yi2016pedestrian}. We scale the coordinates back to the image size of $720 \times 480$ to compute the other two metrics.

\textbf{Results:}
We select the first 80\% of the tracklets as the training set, then 10\% as validation and the last 10\% as test, and report the test error.  
As can be seen from Table \ref{tab:err}, our simple baseline, vanilla LSTM, is on par with Behavior CNN in \cite{yi2016pedestrian}. Although our experimental settings are not identical to \cite{yi2016pedestrian}, this rough comparison still
verifies the effectiveness of our encoder-decoder architecture of the policy. Slight improvement is achieved with SA-LSTMs or collision avoidance respectively, preliminarily implying the effectiveness of the proposed social-aware components.

Our full SA-GAIL without code performs the best, attributed to the better modeling of the decision-making process with collision avoidance and SA-LSTMs for inter-human interaction, as well as the suitable training algorithm, GAIL, for imitating the process. Ablation tests also demonstrate the necessity of each introduced module, though from the comparison between the ablation tests and baselines, not all network architectures benefit from adversarial training, 
presumably due to the difficulty of training recurrent neural network generators with GANs \cite{45829}. The comparison also suggests that the effects of the collision avoidance regularization need further investigation.

\subsection{Intention Inference}

We add the posterior and code for intention inference. 
We experimented with both a 2-dimensional one-hot code and a 3-dimensional one-hot code. We expect one code configuration discovers the latent intention, and the other configuration(s) approximately model behaviors under different crowd scenarios. We approximately treat the intentions in this scene to be ideal straight paths to one of the ten entrances/exits as shown in Fig. \ref{fig:scene}.
After training with SA-GAIL, we generate different future paths for each observed trajectory by iterating over all the code configurations. The results are shown in Fig. \ref{fig:intent}, where (1-6) are results of the 3-dimensional code, and (7-9) are results of the 2-dimensional code.

We managed to achieve that in both experiments one configuration of the latent code would make the policy exhibit a more exit-oriented behavior than the other configuration(s) directly towards the middle of the exits. We arrange the colors of the generated paths so that both such codes in the two experiments correspond to green trajectories, and use green dashed lines to illustrate the straight orientations. With information maximization the generated trajectories don't (and don't have to) mimic the ground-truth very accurately, but at least share similar trend.

We also show some failure cases in (10-12) in Fig. \ref{fig:intent}. Sudden turns after the last observed coordinates inevitably cause large discrepancy as in (10).
Occasional inability to recognize the scene (such as walls in (12)) is also present with no such labels in data. Our policy may also fail on some rare cases. 
Besides, it's hard to evaluate this task fully objectively since it requires manual specification of intention and other criteria.
Still, we demonstrate the potential of our method through direct visualization, a key test as in \cite{chen2016infogan,li2017inferring}.

\section{Conclusions}

In this paper, we propose a novel framework for understanding human behaviors in crowds by inferring their underlying decision-making process, unlike recent advances through supervised deep learning.
Our system, harnessing collision avoidance and other human interaction with recurrent architectures, models the decision-making process and disentangles the different decision factors by introducing modes to the policy.
Experiments demonstrate that our algorithm can not only predict proper future paths of pedestrians but also infer their inner intentions. By understanding and imitating the human decision process, it provides a good opportunity to develop a system that could make safe and reasonable decisions in unconstrained crowded scenarios.

\section*{Acknowledgements}
The first author would personally like to thank his beautiful and sweet wife, Jiamin Deng, for her incredible support during the whole process of this paper.

\bibliographystyle{aaai}
\bibliography{refs}

\end{document}